\documentclass[letterpaper, 10 pt, conference]{ieeeconf} 
\usepackage[letterpaper,top=55pt,bottom=55pt,left=54pt,right=54pt]{geometry}
\usepackage{graphicx}
\usepackage{amsmath}
\usepackage{makecell}
\usepackage{amssymb}
\usepackage{subfigure}
\usepackage{stfloats}
\usepackage{algorithm}
\usepackage{algorithmicx}
\usepackage{algpseudocode}
\usepackage{url}
\usepackage{tabularx, threeparttable,supertabular,ltxtable}
\usepackage{array}
\usepackage{multirow}
\usepackage{color}
\usepackage{url}
\usepackage{amsmath, bm}
\usepackage{hyperref}
\hypersetup{
	colorlinks,
	linkcolor={black},
	citecolor={black},
	urlcolor={black},
	pdfsubject={Physics}
}
\IEEEoverridecommandlockouts 
\title{\LARGE \bf
PriorLane: A Prior Knowledge Enhanced Lane Detection Approach Based on Transformer}

\author{Qibo Qiu$^{1,2}$, Haiming Gao$^{1}$,  Wei Hua$^{1\dagger}$, Gang Huang$^1$ and Xiaofei He$^2$, ~\IEEEmembership{Senior Member, ~IEEE}
\thanks{This work is supported in part by the Key Research and Development Program of Zhejiang Province (No. 2021C01012).}%
\thanks{$^{1}$Zhejiang Lab, Hangzhou, P. R. China, 311121.}
\thanks{$^{2}$State Key Lab of CAD\&CG, College of Computer Science, Zhejiang University, Hangzhou, P. R. China, 310058.}
\thanks{$\dagger$Corresponding author.}
}
\begin{document}
\maketitle
\thispagestyle{empty}
\pagestyle{empty}

\begin{abstract}
Lane detection is one of the fundamental modules in self-driving. In this paper we employ a transformer-only method for lane detection, thus it could benefit from the blooming development of fully vision transformer and achieve the state-of-the-art (SOTA) performance on both CULane and TuSimple benchmarks, by fine-tuning the weight fully pre-trained on  large datasets. More importantly, this paper proposes a novel and general framework called PriorLane, which is used to enhance the segmentation performance of  the fully vision transformer by introducing the low-cost local prior knowledge.  Specifically, PriorLane utilizes an encoder-only transformer to fuse the feature extracted by a pre-trained segmentation model with prior knowledge embeddings. Note that a Knowledge Embedding Alignment (KEA) module is adapted to enhance the fusion performance by aligning the knowledge embedding. Extensive experiments on our Zjlab dataset show that PriorLane outperforms SOTA lane detection methods by a $2.82\%$ mIoU when prior knowledge is employed, and the code will be released at: {\url{https://github.com/vincentqqb/PriorLane}}.

\end{abstract}

\section{Introduction}
Lane detection is one of the most essential tasks of perception for both self-driving and Advanced Driver Assistant System (ADAS), which provides information for localization and planning \cite{kawasaki2020multimodal}. Many traditional lane detection methods are designed based on hand-crafted features \cite{brandes2020robust, beyeler2014vision}. With a blooming development of deep learning, lane detection comes towards a new age with the higher robustness and the stronger generalization, because the lane detection task depends on texture features and high-level semantic analysis, and it benefits a lot from the representation ability of deep learning based models. 

In order to carry out the efficient lane detection task, the segmentation based method, as one of the most typical methods in the field of lane detection, has attracted much attention \cite{icra2019dfnet, abualsaud2021laneaf}, while there are several methods designed to take advantage of the strong shape prior of lane markings, e.g., anchor-based methods \cite{li2019line, tabelini2021keep}, row-wise detection methods \cite{qin2020ultra, philion2019fastdraw}, parametric prediction methods \cite{liu2021end, tabelini2021polylanenet}.

Although those aforementioned methods have made a great progress in the field of autonomous vehicles, the research on robust lane detection is still far from mature. Especially, severe occlusions caused by vehicles and pedestrians are frequently occurring, and lane markings in the distance are too fuzzy to catch the feature in deep models. 

In this paper, a pure vision transformer with a hierarchical encoder is employed to segment lane markings in images, which is called Mixed Transformer (MiT) \cite{xie2021segformer}. Because the transformer is designed to capture long-term dependency, and it has recently demonstrated greater potential than CNN in the rapidly expanding big data scene. In addition, the MiT blocks could be fully pre-trained on large scale datasets, since the transformer-only architecture used in lane detection and general segmentation are unified in an elegant way.

Furthermore, to enhance the segmentation performance, a general fusion framework called PriorLane is proposed to leverage the prior knowledge to improve the performance of MiT blocks. The local prior knowledge data is represented by a grid map in Bird's Eye View (BEV), then PriorLane cuts the grid map into patches and maps them with a trainable linear projection, which are known as embeddings. And a corresponding Knowledge Embedding Alignment (KEA) module is designed to align the embedding spatially.

Finally, a novel fusion transformer (FT) is used to fuse the image feature extracted via MiT blocks with embeddings of the prior knowledge data. Similar to the segformer \cite{xie2021segformer}, an Multi-Layer Perceptron (MLP) block is adapted to merge the fused feature with pure MiT features, and outputs pixel-level results of segmentation prediction.

The contributions of this paper are concluded as follows:
\begin{itemize}
	\item To our knowledge, the transformer-only architecture is employed to perform lane detection for the first time, and it could benefit from the development of fully vision transformer pre-training, which is supported by our experiments on CULane and TuSimple benchmarks. 
	\item More importantly, a novel and general framework called PriorLane is proposed, which enhances the performance of lane segmentation via fusing the image feature with the low-cost local prior knowledge and obtains the better performance by comparative experiments.
	\item The KEA block is further employed for the prior knowledge alignment, and experiments on our Zjlab dataset show that our framework achieves SOTA performance, e.g., an 73.78\% mIoU (2.82\% higher than SOTA).
	
\end{itemize}
\section{Related Work}
\subsection{Segmentation-based Lane Detection}
Different from general segmentation tasks, the structure of lane markings is thin and long, thus it is easy to lose the valid information during feature extracting. To tackle this issue, SCNN \cite{pan2018spatial} proposes a sequential message pass scheme, which is useful for structured objects segmentation. While RESA \cite{zheng2021resa} aggregates global information by shifting the sliced feature map recurrently, and achieves the more efficient performance. Although, segmentation as a kind of pixel-level prediction task suffers from high computation cost and is inflexible for motion planning, it still plays an important role in some applications, e.g., High-Definition map construction and localization \cite{asghar2020vehicle, petek2022robust}.

\begin{figure*}[!b]
	\centering
	\includegraphics[width=18cm]{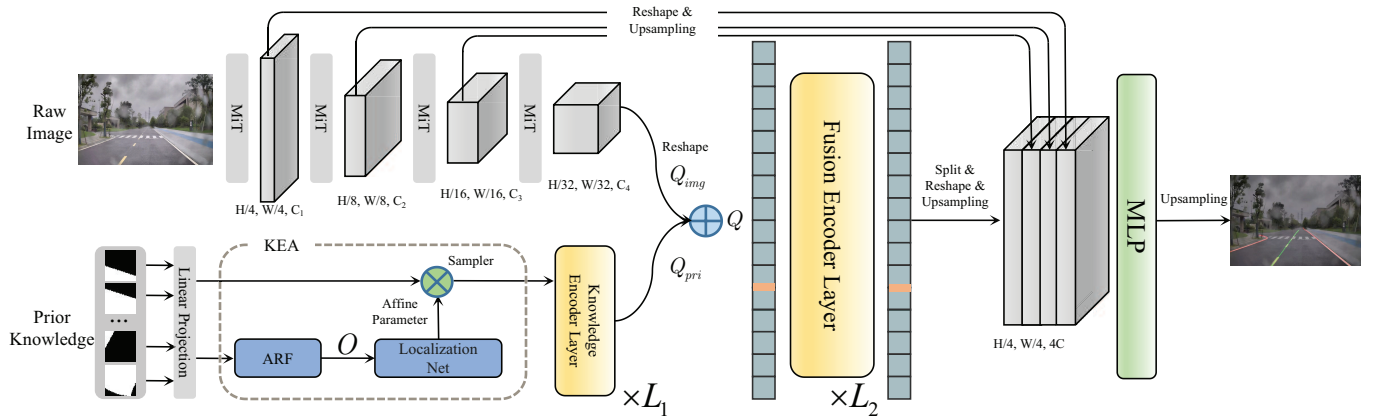}
	\caption{Overview of the proposed PriorLane framework. $L_1$ and $L_2$ indicates the number of knowledge encoder layers and fusion encoder layers, respectively. $O$ is the orientation-invariant feature extracted by ARF.}
	\label{architecture}
\end{figure*}
\subsection{Query-based Object Detection}
DETR \cite{carion2020end}  is the first query-based object detection work which makes use of transformer to decode predictions end-to-end, thus there is no post process, e.g., Non-Maximum Suppression (NMS). In addition, LSTR \cite{liu2021end} achieves a SOTA performance on TuSimple benchmark by regressing the lane line parameters using a Hungarian loss, which is similar to DETR, however the ResNet backbone and transformer module could not be pre-trained simultaneously.

To mitigate the problem, FP-DETR \cite{wang2021fp} treats the query as a kind of implicit prompts, thus both the backbone and attention layers could be pre-trained simultaneously for object detection. Different from FP-DETR, our PriorLane framework makes use of prompts to fine-tune the segmentation model, in particularly, the prompt is explicit and no manually annotated labels are used to train it.

\subsection{Multi-View Transformers}
BEVFormer \cite{li2022bevformer} use the query initialized in BEV to merge the feature from multi camera views and predict corresponding 3D bounding boxes. \cite{saha2022translating} also uses the attention mechanism to map the camera view into the BEV representation. Aforementioned approaches depend on intrinsic and extrinsic calibration of cameras to build the relationship between BEV and camera views, moreover the BEV representation is usually initialized randomly and can be learned during training. The query used in the proposed PriorLane framework has a physical meaning, and corresponding intrinsic and extrinsic parameters are not needed, while high precision positions and orientations are expensive to acquire in real applications.
\section{Method}
\subsection{Architecture Design}
Fig. \ref{architecture} shows the overall architecture of the proposed PriorLane framework, it mainly consists of three successive parts: the MiT block, the KEA module and the FT block.

\textbf{Mix Transformer.} \cite{xie2021segformer} designs a series of MiT blocks, and MiT-B5 is utilized in this paper, which is the largest version of MiT blocks and has the best performance on segmentation. Different from typical ViT \cite{dosovitskiy2020vit}, the advantage of MiT could generate hierarchical features, due to the design of overlapped patch merging and efficient self-attention.

\textbf{Knowledge Embedding Alignment. }Since the vehicle positions are coarse, it is hard to determine the reference knowledge embedding in BEV consistent with the camera view. To address this issue, a Knowledge Embedding Alignment(KEA) module is needed to align the knowledge embedding with image features spatially.

\textbf{Fusion Transformer.} 
The knowledge encoder layer and the fusion encoder layer work together as the fusion transformer. Different from FP-DETR, which proposes a novel approach for detection transformer pre-training by taking advantage of implicit vision prompts. In this paper, prompts in PriorLane are generated by explicit prior knowledge, moreover, no annotated labels are used for prompts training.

\begin{figure}[htp]
	\centering
	\includegraphics[width=6cm]{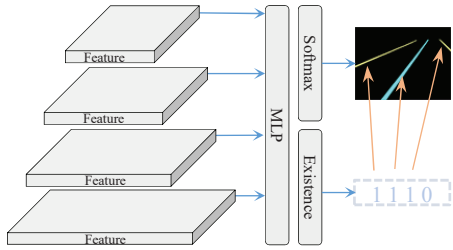}
	\caption{The segmentation architecture named MiT-Lane on CULane and TuSimple benchmarks, and 0/1 indicates the existence of the corresponding lane marking.}
	\label{bench_arch}
\end{figure}

\textbf{Architecture on benchmarks.} Typical lane segmentation methods on CULane and TuSimple benchmarks leverage the annotation of lane existence, e.g., RESA, SCNN and UFLD \cite{qin2020ultra}. To verify the performance of fully vision transformer on CULane and TuSimple benchmarks, the MiT-Lane is employed, which is an intuitive extension of segformer by adding a bench for existence prediction following the MLP block, as shown in Fig. \ref{bench_arch}.

\subsection{Local Prior Knowledge Representation} 
\begin{figure}[htp]
	\centering
	\includegraphics[width=6cm]{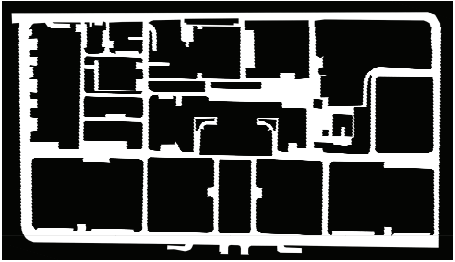}
	\caption{The prior knowledge of Zhejiang Lab park represented as a grid map.}
	\label{zjmap_binary}
\end{figure}

Road network, elevation map, etc., are known as the prior knowledge of traffic scenes. In this paper, the prior knowledge is stored in BEV by a grid map, and Fig. \ref{zjmap_binary} shows a grid map with one channel, while 0/1 indicates whether the corresponding grid is in the road freespace or not. More importantly, other kinds of prior knowledge could also be leveraged by adding additional channels. In this way the prior knowledge is rendered into a ``big image'' and the local prior knowledge could be represented as a smaller one once the perception range is given.

The vehicle positions are collected by a meter-level equipment, and the orientation of the vehicle is randomly rotated to simulate a coarse position, since it is high-cost to obtain the accurate position in an urban environment. And PriorLane is proposed to enhance the lane segmentation performance without high-definition prior knowledge. Once the resolution, the perception range and the position are given, the sub-image from the ``big image" could be obtained, which is named as ``prior knowledge" hereinafter.

\textbf{Knowledge Embedding.} Due to the powerful ability of feature representation by transformer, only  linear projection is needed for knowledge embedding. Firstly, we cut the local grid map data $M \in \mathbb{R}^{H \times W \times C}$ into patches and the size of each patch is $P \times P$. Moreover, each patch is converted into a vector of size $1 \times E_p$ by a trainable linear projection, thus the local prior knowledge is represented as the ``knowledge embedding", which is denoted by $X$ and the corresponding size is $(H/P)\times (W/P) \times E_p$.

\subsection{Knowledge Embedding Alignment} 
In this section, Spatial Transformer Network (STN) \cite{jaderberg2015spatial} is adapted to align the embedding in the KEA module. STN could manipulate the prior knowledge embedding according to the affine transform matrix obtained by a localization network, thus the fusion performance of the prior knowledge and the image feature could be enhanced.

Traditional CNN lacks the capability to extract direction information, which is crucial for STN. Active Rotation Filters (ARFs) and ORPooling  \cite{zhou2017oriented, han2021align} are adapted to extract orientation information from the prior knowledge embeddings and keep the rotation-invariant information, respectively.
An ARF $R$ is made up of $N$ canonical filters, whose size is $N \times K \times K$, and $N$ means rotating the kernel of each filter for $N-1$ times, the degree is $\frac{2 \pi}{N}$ per time. For a given feature map $X$, the output of an ARF is $F$ denoted as
\begin{equation}{F}^{(i)}=\sum_{n=0}^{N-1}{R}_{\theta_{i}}^{(n)} \cdot{X}^{(n)}, \theta_{i}=i \frac{2 \pi}{N}, i=0, \ldots, N-1, \end{equation}
where ${F}^{(i)}$ is the $i$-th orientation of $F$, and ${R}_{\theta_{i}}$ means rotating $R$ by a degree of $\theta_{i}=i \frac{2 \pi}{N}$, ${R}_{\theta_{i}}^{(n)}$ indicates the $n$-th channel of $R_{\theta_{i}}$ and ${X}^{(n)}$ is the $n$-th orientation channel of the corresponding feature map $X$. We pool the orientation-sensitive feature $F$ and keep rotation-invariant information $O$ as follows
\begin{equation}O = max\ {F}^{(n)}, 0 \leq n < N,\end{equation}
where ${F}^{(n)}$ indicates the $n$-th orientation channel of $F$. The rotation-invariant feature $O$ is only used for projection parameters estimation, and the output after affine transformation is sampled from the original embedding of the local prior knowledge $X$, as shown in Fig. \ref{architecture}. 
\subsection{Fusion Transformer Layers} 
As shown in Fig. \ref{architecture}, PriorLane uses knowledge encoder layers with self-attention to refine the embedding aligned by the KEA module. The output of knowledge encoder layers can be treated as a kind of prompts for fusion encoder layers, and inter-prompt relationships can be learned in the fusion encoder layer. 
As described in \cite{meng2021conditional}, the cross-attention usually results in a slow convergence, therefore self-attention is utilized to fuse the prior knowledge embedding with image features in fusion encoder layers, and it is calculated as
\begin{equation}\operatorname{Attention}(Q, K, V)=\operatorname{Softmax}\left(\frac{Q K^{\top}}{\sqrt{d_{k}}}\right) V,\end{equation}
where
\begin{equation}Q = K = V = [Q_{img}; Q_{pri}].\end{equation}
$Q$ is the concatenation of $Q_{img}$ and $Q_{pri}$, and $Q_{pri}$ is the output of the knowledge encoder layer with the size of $L \times E_p$, while $L=(H/P)\times (W/P)$. Considering that images are collected by different cameras, which have different resolutions, intrinsic parameters and extrinsic parameters, position encoding is  not employed in attention calculation.

\section{Experiments}
\subsection{Datasets}
In this paper, we firstly conduct experiments on two famous benchmarks: CULane \cite{pan2018spatial} and TuSimple\footnote{\textcolor[rgb]{0,0,0}{\url{https://github.com/TuSimple/tusimple-benchmark/}}}. Since there is no public dataset contains images, labelled lane markings  and prior knowledge,  a dataset is collected during the daily operation of unmanned vehicles in Zhejiang Lab park, which is named Zjlab dataset in Table~\ref{ld dataset}. 

\textbf{CULane.} For CULane, F1-measure is adapted for evaluation, similar to \cite{pan2018spatial}, each lane marking is treated as a line with 30 pixel width, thus intersection-over-union (IoU) could be calculated between the ground truth and the predicted segmentation. Predictions with IoUs larger than 0.5 are treated as true positives (TP) in our experiments. The F1-measure is defined as: $F_1 = \frac{Precision  Recall}{Precision+Recall}$, where $Precision = \frac{TP}{TP+FP}$ and $Recall = \frac{TP}{TP+FN}$, $FP$ means false positive and $FN$ indicates false negative.

\begin{table*}[!t]
	\center
	\caption{Lane detection datasets description}
	\renewcommand{\arraystretch}{1.4}
	\begin{tabular}{cccccccc}
		\hline \hline
Dataset  & Train  & Test &Visualize&Resolution& Lane & Environment & Labels' Meaning        \\ \hline
		TuSimple & 3268   & 2782 & 2782 & $720\times1280$ &$\leq 5$   & highway       & from left to righe     \\ \hline
		CULane   & 133235 & 34680 & 34680 & $590\times1640$ & $\leq 4$   & urban\textbackslash{}highway & from left to right     \\ \hline
		Zjlab    & 695    & 174   &  $10000$& $1080\times1920$& $\leq 8$   & park          & yellow\textbackslash{}white\textbackslash{}stop line \\ \hline \hline
	\end{tabular}
	\label{ld dataset}
\end{table*}

\textbf{TuSimple.} There is an official accuracy evaluation metric for TuSimple, which is defined as: $accuracy = \frac{\sum_{clip} C_{clip}}{\sum_{clip} S_{clip}}$, where $C_{clip}$ is the number of lane key points that predicted correctly and $S_{clip}$ is the number of ground truth lane key points in each clip (only the last frame of each clip is under evaluation). A lane key point is viewed as a true positive only if its distance from the ground truth is within 20 pixels.

\textbf{Zjlab.} To collect Zjlab dataset, we mounted cameras with a resolution of 1920 $\times$ 1080 on 3 vehicles during the daily operation. Finally, 869 typical frames with different vehicle positions and weathers are collected and labelled, e.g., sunny, cloudy and rainy. There are also bends, straights and junctions with lane markings  blocked by obstacles. The labelled dataset is shuffled and divided into train and test sets, which have 695 and 174 images respectively. In addition, 10000 images are collected for visualization.

To access the semantic segmentation performance, the mean intersection-over-union (mIoU) is calculated following \cite{cordts2016cityscapes}, which averages over IoUs of all classes, and the IoU is defined as: $IoU = \frac{TP}{TP+FP+FN}$, where $TP$, $FP$ and $FN$ indicate the number of true positive, false positive and false negative pixels. A pixel is treated as a true positive only when the predicted label is equal to the ground truth.

\subsection{Experiment Settings}
For Zjlab dataset, three versions of PriorLane are carried out: PriorLane-Imp, PriorLane-KE and PriorLane-KEA. Experiments of PriorLane-KE are conducted to verify the influence of prior knowledge embeddings. Moreover, PriorLane-KEA is carried out by adding the KEA module to PriorLane-KE. In addition, PriorLane-Imp is designed, whose architecture is the same as PriorLane-KEA, while trainable weights with randomly initialization are put into the KEA module instead of the knowledge embedding. 

The RESA and SCNN models on Zjlab dataset are implemented base on the pytorch-auto-drive \cite{feng2022rethinking} codebase and the detailed configurations stay the same with those on CULane benchmark published by the codebase. Models on Zjlab dataset are trained on the same server with 8 Tesla V100s and the input size is $540 \times 960$, and the main configurations of PriorLane models have no difference between those on CULane and TuSimple benchmarks. 

\subsection{Main Results}

\begin{figure*}[htp]
	\center
	\includegraphics[width=17cm]{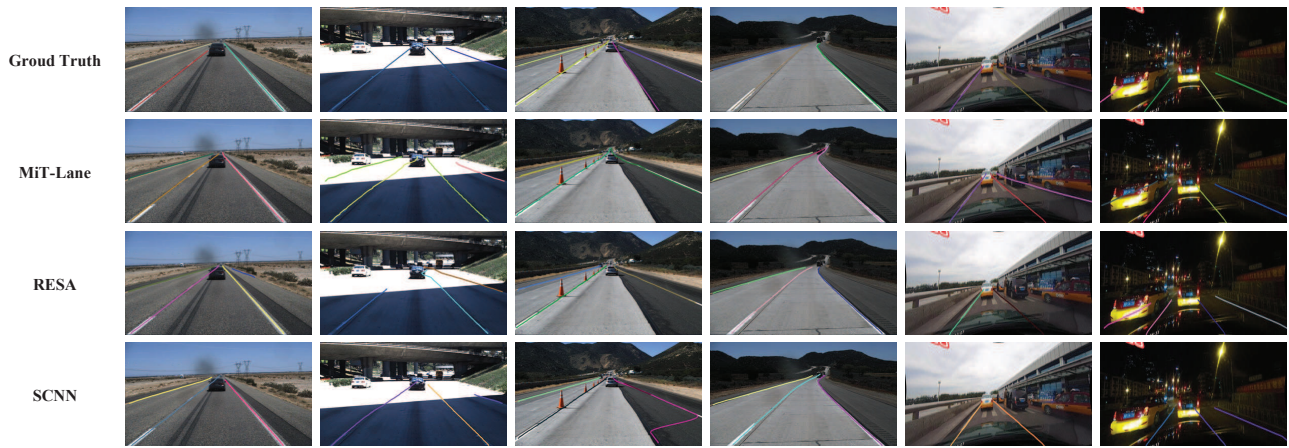}
	\caption{Visualization results on CULane and TuSimple datasets, the first four columns are results on the TuSimple dataset, while the rest columns are results on the CULane dataset. Different lane marking instances are represented by different colors, note that MiT-Lane can predict a longer lane marking compared with the ground truth, however the longer part negatively affects the quantitative results.}
	\label{tusimple_vis}
\end{figure*}

\textbf{Results on CULane. }To compare the results on CULane dataset with SOTA methods, four popular segmentation-based methods are used for comparison, including SCNN, UFLD and RESA. Although UFLD is often classified as a row-wise method, compared with anchor-based methods it is more similar to the segmentation method, and auxiliary segmentation is used during training. Table \ref{culane_table} shows the F1-measures on the CULane dataset, and the modified method MiT-Lane outperforms other segmentation-based methods. In Fig. \ref{tusimple_vis}, qualitative results on CULane dataset show that MiT-Lane based on fully vision transformer can predict the lanes occluded severely, because transformer can not only catch the longitudinal dependency, but also the inter-relationship between different lane markings.

\textbf{Results on TuSimple.} For TuSimple benchmark, LSTR is also used for comparison, because it is an early transformer-based method in lane detection. Table. \ref{tusimple_table} shows the quantitative results on TuSimple benchmark, and MiT-Lane achieves the SOTA performance. Visualization results on TuSimple benchmark are shown in Fig. \ref{tusimple_vis}, and the MiT-Lane method can predict a longer lane marking based on the long-term dependency modeling ability of transformer, which is important for the highway scenario. However, the longer part negatively affects the quantitative results, because it has no human-annotated labels.

\begin{figure*}[htp]
	\centering
	\includegraphics[width=17cm]{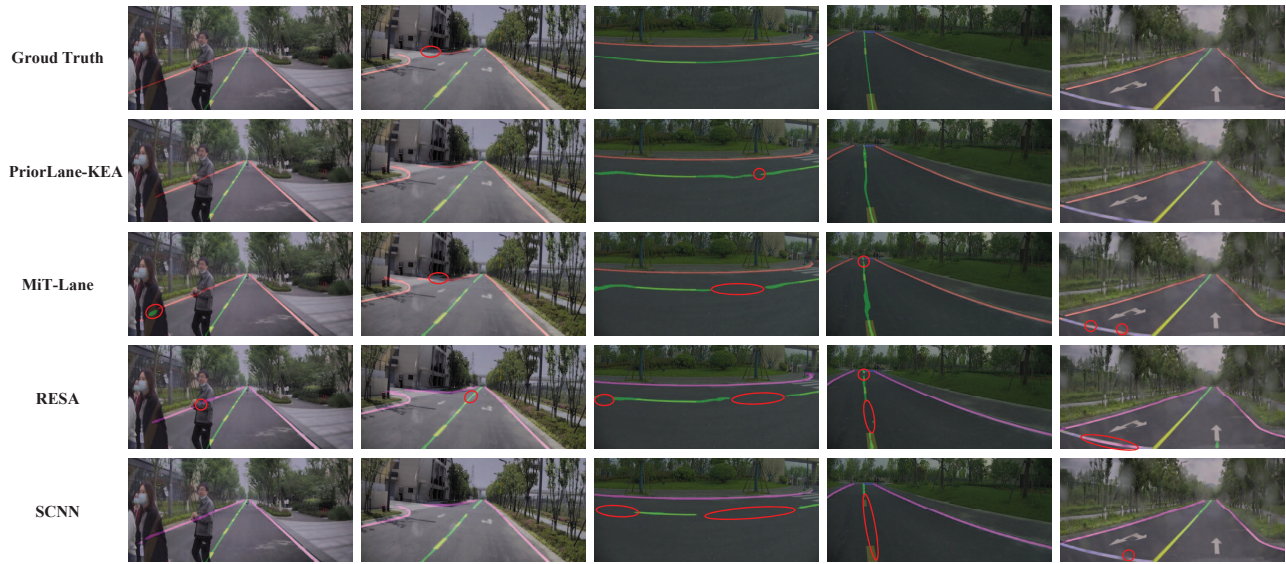}
	\caption{Visualization results on Zjlab dataset, different types of lane markings are represented by different colors, and red circles indicate failures of ground truths and predictions. Note that PriorLane-KEA can predict the blank of dotted line markings, because of the ability of long-term dependency modeling, and the prior knowledge also contributes to pixel semantic classification.}
	\label{zjlab_vis}
\end{figure*}

\textbf{Results on Zjlab.} For Zjlab dataset, SCNN, RESA, MiT-Lane are used as baselines. Quantitative results in Table \ref{zjlab_res} show that the proposed framework PriorLane-KEA get the best mIoU of 73.78\%, which indicates that prior knowledge can improve the segmentation performance. Visualization results in Fig. \ref{zjlab_vis} show that PriorLane-KEA can predict the blank of dotted line markings and the blocked part more accurately, which benefits from the local prior knowledge. In addition, the model can distinguish stop lines from white lane markings at a intersection.  

Comparative experiments show that MiT-Lane outperforms SCNN and RESA on TuSimple, CULane and Zjlab datasets, which verifies that the transformer-only lane segmentation method without domain specific design is a qualified lane detector. Thus, lane segmentation could benefit a lot from the fast-growing progress of the fully vision transformer, which is verified to be powerful on large datasets.

\begin{table*}[!t]
	\centering
	\caption{Comparison of F1-measure with SOTA methods on CULane test set, IoU threshold is setted 0.5, and only FP is shown for crossroad. * reproduced by the pytorch-auto-drive framework, which is the best result of 3 random runs, ** results in official paper. The performance of MiT-Lane is the mean value from 3 random runs.}
	\renewcommand{\arraystretch}{1.3}
	\begin{tabular}{lcccccccccc}
		\hline \hline
		Methods & Total & Normal & Crowed & Night & No line & Shadow & Arrow & Dazzle & Curve & Cross \\ \hline
		SCNN(ResNet-101)* & 73.58 & 91.10 & 71.43 & 68.53 & 46.39 & 72.61 & 86.87 & 61.95 & 67.01 & 1720 \\ \hline
		UFLD (ResNet-34)** & 72.3 & 90.7 & 70.2 & 66.7 & 44.4 & 69.3 & 85.7 & 59.5 & 69.5 & 2037 \\ \hline
		RESA (ResNet-50)* & 74.04 & 91.45 & 71.51 & 69.01 & 46.54 & 75.83 & 87.75 & 63.90 & 68.24 & \textbf{1522} \\ \hline
		MiT-Lane & \textbf{76.27} & \textbf{92.36} & \textbf{73.86} & \textbf{70.26} & \textbf{49.6} & \textbf{78.13} & \textbf{88.59} & \textbf{68.26} & \textbf{73.94} & 2688.3 \\ \hline \hline 
	\end{tabular}
	\label{culane_table}
\end{table*}

\begin{table}[!t]
	\centering
	\caption{Comparsion of Accuracy, FP, FN on TuSimple test set, the max length of the prediction is setted 56. * reproduced by the pytorch-auto-drive framework, which is the best result of 3 random runs, ** results in official paper. The performance of MiT-Lane is the mean value from 3 random runs.}
	\renewcommand{\arraystretch}{1.3}
	\begin{tabular}{llll}
		\hline \hline
		Method & Accuracy & FP & FN \\ \hline
		SCNN(ResNet-101)* & 95.69 & 0.052 & 0.050 \\ \hline
		UFLD(ResNet-34)** & 96.06 & - & - \\ \hline
		RESA(ResNet-101)* & 95.56 & 0.058 & 0.051 \\ \hline
		LSTR(ResNet-18)* & 95.06 & 0.049 & 0.042 \\ \hline
		MiT-Lane & \textbf{96.58} & \textbf{0.039} & \textbf{0.029} \\ \hline \hline
	\end{tabular}
	\label{tusimple_table}
\end{table}

\begin{table}[h!t]
	\centering
	\tabcolsep=3pt
	\caption{Comparison of the IoU on Zjlab test set, each result is the mean value from 3 random runs}
	\renewcommand{\arraystretch}{1.3}
	\begin{tabular}{ll<{\raggedright}m{1cm}<{\raggedright}m{1cm}<{\raggedright}m{1cm}<{\raggedright}m{1cm}<{\raggedright}}
		\hline \hline
		Method & \makecell[c]{Mean \\IoU} & Back- ground & White Line & Yellow Line & Stop Line \\ \cline{1-6}
		SCNN & 67.89 & 99.30 & 66.66 & 51.39 & 54.23 \\ \cline{1-6}
		RESA & 70.96 & 99.43 & 71.88 & 61.12 & 51.41 \\ \cline{1-6}
		MiT-Lane & 72.16 & 99.56 & 74.01 & 61.99 & 53.08 \\ \cline{1-6}
		PriorLane-Imp & 72.62 & 99.57 & 74.95 & 62.63 & 53.31 \\ \cline{1-6}
		PriorLane-KE & 73.28 & 99.58 & \textbf{75.25} & 62.85 & 55.42 \\ \cline{1-6}
		PriorLane-KEA & \textbf{73.78} & \textbf{99.59} & 75.01 & \textbf{63.12} & \textbf{57.39} \\ \hline \hline
	\end{tabular}
	\label{zjlab_res}
\end{table}

\subsection{Ablation Study}
\textbf{Analysis for Components. }PriorLane-KE surpasses MiT-Lane by a 1.12\% mIoU, which indicates the explicit prompts could improve the performance of lane segmentation. Further comparison with PriorLane-KEA shows that the KEA module can improve the result by aligning the local prior knowledge embeddings, supported by a 0.5\% mIoU. Note that PriorLane-Imp, which has the same architecture with PriorLane-KEA, outperforms MiT-Lane by a 0.46\% mIoU, it is mainly due to the effect of the trainable prompts. 

\textbf{Number of Encoder Layers.} There are two kinds of encoder layers: the knowledge encoder layer and the fusion encoder layer, and the number of knowledge encoder layers and fusion encoder layers are indicated by $L_1$ and $L_2$, respectively, as shown in Fig. \ref{architecture}. 

\begin{figure}[htp]
	\centering
	\includegraphics[width=7.5cm]{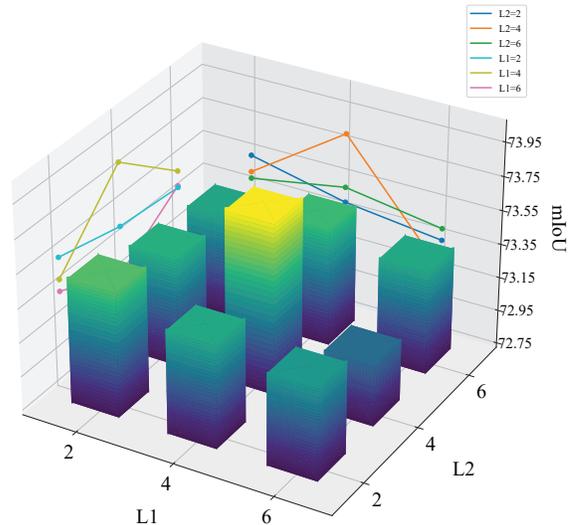}
	\caption{Quantitative results of different settings of $L_1$ and $L_2$, represented by different pillars, and the number of attention heads is fixed to 8. Each result is the mean value from 3 random runs.}
	\label{num_enc_layers}
\end{figure}

To investigate the effect of the number of different encoder layers, experiments with different $L_1$ and $L_2$ values are carried out. As shown in Fig. \ref{num_enc_layers}, mIoUs vary with different $L_1$ and $L_2$ values, and there is a peak in the histogram. The knowledge encoder layer can be treated as a feature refiner, and deeper layers produce higher semantic features, which contribute to the segmentation performance. However, the larger $L_1$ degrades the model generalization and leads to a decrease of the mIoU after the peak. 

Higher semantic features produced by the knowledge encoder layer need deeper fusion encoder layers to merge the valid information. However, the image feature dominates others in deeper fusion encoder layers, thus the performance could be suppressed, as a result,  a larger $L_2$ does not necessarily obtain a higher mIoU under a fixed $L_1$. 

\begin{figure}[htp]
	\centering
	\includegraphics[width=7.5cm]{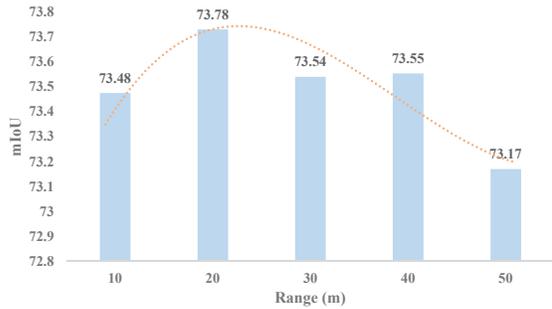}
	\caption{Quantitative results of different perception ranges (the blue histogram), and the orange trend line indicates a downtrend when the perception range is larger than 20m. The value of $L_1$ and $L_2$ are fixed to 4 and 4, respectively, and the number of attention heads is fixed to 8. Each result is the mean value from 3 random runs.}
	\label{zjlab_dist}
\end{figure}

\textbf{Influence of Perception Range. }To investigate the effect of different perception ranges, experiments with different perception ranges are carried out, as shown in Fig. \ref{zjlab_dist}. Since the local prior knowledge is represented in a grid map, which is rendered into an ``image'' with one channel,  and the longer perception range results in a larger ``image". Accordingly, local prior knowledge with different perception ranges are resized to the same size ($200 \times 200$). As shown in Fig. \ref{zjlab_dist}, with the increase of the perception range, PriorLane-KEA obtains a higher mIoU with the help of the local prior knowledge. However, the image resolution in the distance is low, and it is difficult for the model to calculate the attention between the low resolution feature and the corresponding prior knowledge, as a result, there is a decrease when the perception range is too large.

\section{Conclusion}
In this paper, we have verified that the transformer-only architecture on lane
 detection and general segmentation can be unified in an elegant way. Thus, lane detection task could benefit from the rapid development of vision transformer pre-training. In addition, a novel framework called PriorLane is proposed, which fuses the local prior knowledge of the scene with images. Moreover, a KEA module is adapted to align the data from two different modalities, and experiments show that both the framework and KEA module could enhance the performance of lane segmentation. In the future, a larger dataset with more kinds of low-cost prior knowledge will be collected to enhance the segmentation performance.

\end{document}